\documentclass[11pt,a4paper]{article}
\usepackage[hyperref]{acl2021}
\usepackage{times}
\usepackage{latexsym}
\usepackage{graphicx}
\usepackage{amsmath}
\usepackage{amsfonts}
\usepackage{amssymb}
\usepackage{multirow}
\usepackage{makecell}
\usepackage{booktabs}
\usepackage{verbatim}
\usepackage[toc,page]{appendix}

\usepackage{microtype}

\aclfinalcopy 

\title{Comprehensive Study: How the Context Information of Different Granularity Affects Dialogue State Tracking?}

\author{Puhai Yang \and Heyan Huang\thanks{ { } Corresponding author} \and Xian-Ling Mao \\
  School of Computer Science and Technology, \\
  Beijing Institute of Technology, Beijing, China \\
  Beijing Engineering Research Center of High Volume Language \\
  Information Processing and Cloud Computing Applications, Beijing, China \\
  Southeast Academy of Information Technology, Beijing Institute of Technology, Fujian, China \\
  \texttt{\{phyang, hhy63, maoxl\}@bit.edu.cn}}

\date{}

\begin{document}

\maketitle
\begin{abstract}
Dialogue state tracking (DST) plays a key role in task-oriented dialogue systems to monitor the user's goal. In general, there are two strategies to track a dialogue state: predicting it from scratch and updating it from previous state. The scratch-based strategy obtains each slot value by inquiring all the dialogue history, and the previous-based strategy relies on the current turn dialogue to update the previous dialogue state. However, it is hard for the scratch-based strategy to correctly track short-dependency dialogue state because of noise; meanwhile, the previous-based strategy is not very useful for long-dependency dialogue state tracking. Obviously, it plays different roles for the context information of different granularity to track different kinds of dialogue states. Thus, in this paper, we will study and discuss how the context information of different granularity affects dialogue state tracking. First, we explore how greatly different granularities affect dialogue state tracking. Then, we further discuss how to combine multiple granularities for dialogue state tracking. Finally, we apply the findings about context granularity to few-shot learning scenario.
Besides, we have publicly released all codes.

\end{abstract}

\section{Introduction}

\begin{figure}[!t]
	\centering
	\includegraphics[width=\linewidth]{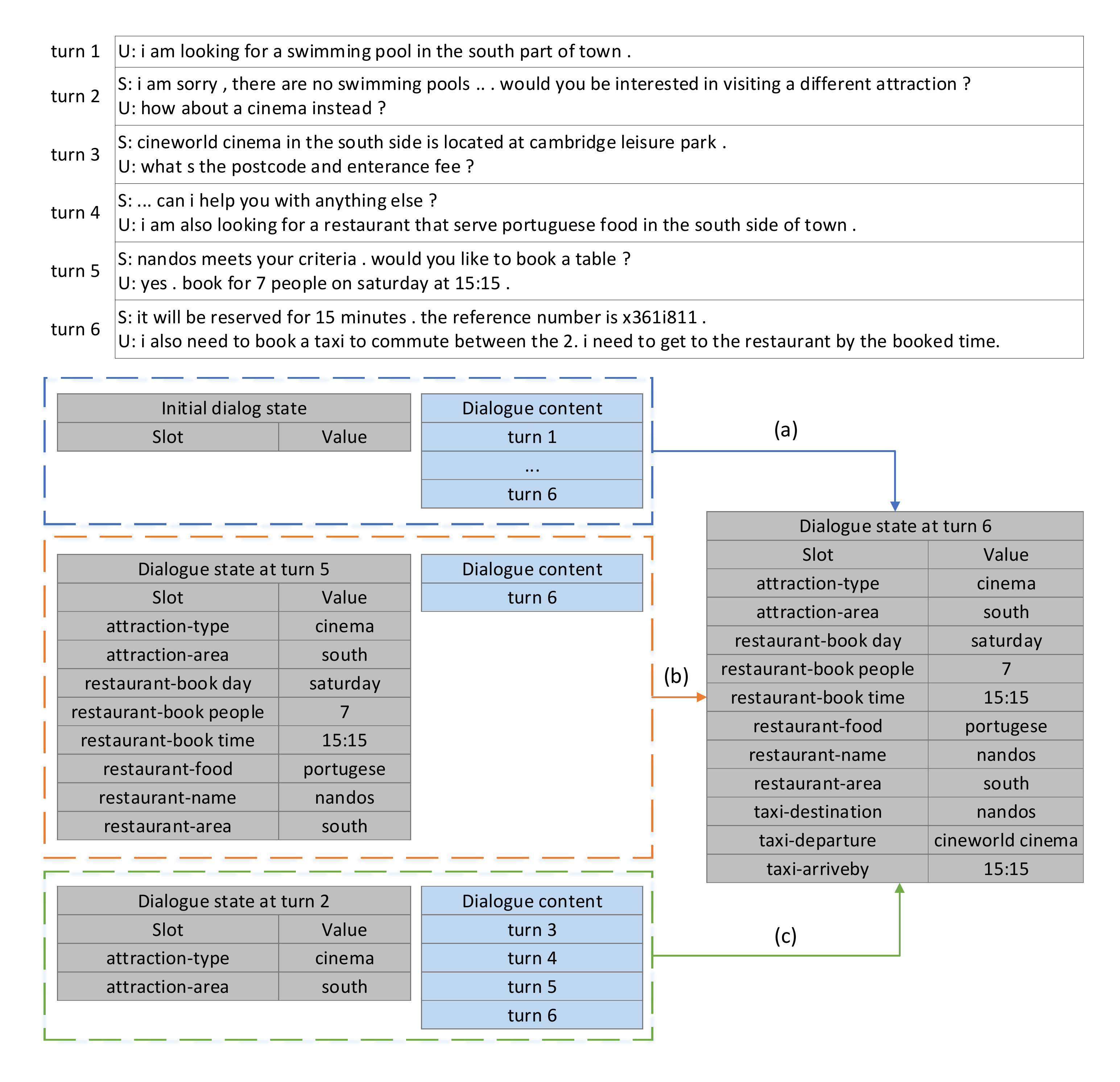}
	\caption{Examples of dialogue state tracking with context information of different granularity at the sixth turn of a dialogue. $Slot$ in a dialogue state refers to the concatenation of a domain name and a slot name. In the figure, (a) represents predicting the dialogue state from scratch, where slots in three domains need to be predicted and the challenge of encoding longer text is faced; (b) indicates updating dialogue state from the previous state, the slot $taxi-departure$ cannot be predicted due to the absence of corresponding dialogue history content; (c) represents dialogue state tracking with context information of granularity 4, which tracks from the second turn and uses less dialogue history content (4 turns) to provide evidence for the prediction of all slots.}
	\label{problem}
\end{figure}

Currently, task-oriented dialogue systems have attracted great attention in academia and industry \cite{chen2017survey}, which aim to assist the user to complete certain tasks, such as buying products, booking a restaurant, etc. As a key component of task-oriented dialogue system, dialogue state tracking plays a important role in understanding the natural language given by the user and expressing it as a certain dialogue state \cite{rastogi2017scalable,rastogi2018multi,goel2018flexible}. The dialogue state for each turn of a dialogue is typically presented as a series of slot value pairs that represent information about the user's goal up to the current turn. For example, in Figure \ref{problem}, the dialogue state at turn 2 is \{($attraction-type$, $cinema$), ($attraction-area$, $south$)\}.

In general, there are two strategies to track a dialogue state: predicting it from scratch and updating it from previous state. The scratch-based strategy obtains each slot value in dialogue state by inquiring all the dialogue history \cite{xu2018end,lei2018sequicity,goel2019hyst,ren2019scalable,wu-etal-2019-transferable,shan2020contextual,zhang2020find}, the advantage of this strategy is to ensure the integrity of the dialogue information. The previous-based strategy relies on the current turn dialogue to update the previous dialogue state \cite{mrkvsic2017neural,chao2019bert,kim-etal-2020-efficient,heck2020trippy,zhu2020efficient}, the main character of this strategy is to greatly improve the efficiency of dialogue state prediction and avoid the computational cost of encoding all dialogue history.

However, both kinds of strategies above have great defects because of their own characters. For the scratch-based strategy, it is hard to correctly track short-dependency dialogue state because of the noise associated with encoding all dialogue history. For example, the dialogue history of turn 1 to 3 in Figure \ref{problem} (a) does not contribute to the prediction of slot values in the $restaurant$ domain. For the previous-based strategy, it is difficult to solve the problem of long-dependency dialogue state tracking because it utilizes only limited dialogue information from the current turn dialogue and the previous state. As in Figure \ref{problem} (b), the slot $taxi-departure$ cannot be predicted due to the absence of corresponding dialogue history content.

Obviously, it plays different roles for the context information of different granularity to track different kinds of dialogue states. Intuitively, less context information is needed for short-dependency dialogue state, while more context information must be taken into account for long-dependency dialogue state tracking. For example, the dialogue state in Figure \ref{problem} (c) is tracked from turn 2, which utilizes context information of granularity 4 (turn 3 to 6), providing evidence for the prediction of all slots while bringing as little noise as possible.

Thus, in this paper, we will study and discuss how the context information of different granularity affects dialogue state tracking. The contribution of this paper is that it is, to the best of our knowledge, the first detailed investigation of the impact of context granularity in dialogue state tracking and promotes the research on dialogue state tracking strategy. Our investigation mainly focuses on three points\footnote{The code is released at \url{https://github.com/yangpuhai/Granularity-in-DST}}:

\begin{itemize}
	\item How greatly different granularities affect dialogue state tracking?
	\item How to combine multiple granularities for dialogue state tracking?
	\item Application of context information granularity in few-shot learning scenario.
\end{itemize}

The rest of paper is organized as follows: The relevant definitions and formulas in the dialogue state tracking strategy are introduced in section 2. Section 3 lists the detailed experimental settings. Section 4 presents the survey report and results, followed by conclusions in section 5.

\begin{table*}[!t]
	\centering
	\begin{tabular}{cccccccccc}
		\hline
		\multirow{2}{*}{Dataset} & \multirow{2}{*}{\# Domains} & \multirow{2}{*}{\# Slots} & \multirow{2}{*}{Avg. turns} & \multicolumn{3}{c}{\# Dialogues} & \multicolumn{3}{c}{\# Turns} \\
		& & & &  train & dev & test & train & dev & test \\
		\hline
		Sim-M & 1 & 5 & 5.14 & 384 & 120 & 264 & 1,973 & 627 & 1,364  \\
		Sim-R & 1 & 9 & 5.53 & 1,116 & 349 & 775 & 6,175 & 1,489 & 3,436  \\
		WOZ2.0 & 1 & 3 & 4.23 & 600 & 200 & 400 & 2,536 & 830 & 1,646  \\
		DSTC2 & 1 & 3 & 7.24 & 1,612 & 506 & 1,117 & 11,677 & 3,934 & 9,890  \\
		MultiWOZ2.1 & 5 & 30 & 6.53 & 8,420 & 1,000 & 999 & 54,984 & 7,371 & 7,368  \\
		\hline
	\end{tabular}
	\caption{Data statistics of Sim-M, Sim-R, WOZ2.0, DSTC2 and MultiWOZ2.1. $Avg.\ turns$ indicates the average number of turns involved in the dialogue in the training data.}
	\label{datasets}
\end{table*}

\begin{table*}[!t]
	\centering
	\begin{tabular}{cccccc}
		\hline
		Models & {Open vocabulary} & Encoder & Decoder & {Tracking strategy} \\
		\hline
		SpanPtr \cite{xu2018end} & \checkmark & RNN & Extractive & {scratch-based} \\
		TRADE \cite{wu-etal-2019-transferable} & \checkmark & RNN & Generative & {scratch-based} \\
		BERTDST \cite{chao2019bert} & \checkmark & BERT & Extractive & previous-based \\
		SOMDST \cite{kim-etal-2020-efficient} & \checkmark & BERT & Generative & previous-based \\
		SUMBT \cite{lee2019sumbt} & $\times$ & BERT & Classification & previous-based \\
		\hline
	\end{tabular}
	\caption{Statistics on the characteristics of the 5 baselines studied in the paper. In the decoder, the extractive mode refers to the extraction of slot values directly from the dialogue context, the generative mode refers to the vocabulary-dependent sequence decoding, and the classification mode is the slot value ontology-based classification.}
	\label{baselines}
\end{table*}

\section{Preliminary}

To describe the dialogue state tracking strategy, let's introduce the formula definitions used in this paper:

\paragraph{Dialogue Content:}
$D=(T_1,T_2,...,T_N)$ is defined as the dialogue of length $N$, where $T_i=(S_i,U_i)$ is the dialogue content of $i$-th turn, which includes the system utterance $S_i$ and the user utterance $U_i$.
\paragraph{Dialogue State:}
We define $E=(B_0,B_1,B_2,$ $...,B_N)$ as all dialogue states up to the $N$-th turn of the dialogue, where $B_i$ is the set of slot value pairs representing the information provided by the user up to the $i$-th turn. In particular, $B_0$ is the initial dialogue state which is an empty set.
\paragraph{Granularity:}
In dialogue state tracking, the number of dialogue turns spanning from a certain dialogue state $B_m$ in the dialogue to the current dialogue state $B_n$ is called granularity, that is, $G=|(T_{m+1},...,T_n)|$. For example, the granularities of context information in (a), (b), and (c) in Figure \ref{problem} are 6, 1, and 4, respectively.

Assuming that the dialogue state of the $N$-th turn is currently required to be inferred, the dialogue state tracking under a certain granularity is as follows:
$$B_N=tracker((T_{N-G+1},...,T_N),B_{N-G})$$
where $G\in\{1,2,...,N\}$ is the granularity of context information and $tracker$ represents a dialogue state tracking model.

In particular, if $G=1$, then:
$$B_N=tracker(T_N,B_{N-1})$$
this case corresponds to the strategy of updating from previous state. Therefore, the previous-based strategy is a special case where context granularity is minimal in dialogue state tracking.

If $G=N$, then:
$$B_N=tracker((T_1,...,T_N),B_0)$$
this case corresponds to the strategy of predicting state from scratch. Similarly, the scratch-based strategy is also a special case of dialogue state tracking, with the context information of maximum granularity. Since the size of the maximum granularity $N$ is different in different dialogues, so 0 is used in the paper to refer to the maximum granularity $N$, -1 to refer to granularity $N-1$, and so on.

\section{Experimental Settings}
In order to investigate how the context information of different granularity affects dialogue state tracking, we analyze the performance of several different types of dialogue state tracking models on different datasets. For a clearer illustration, the detailed settings are introduced in this section.

\subsection{Datasets}
Our experiments were carried out on 5 datasets, Sim-M \cite{shah2018building}, Sim-R \cite{shah2018building}, WOZ2.0 \cite{wen2016network}, DSTC2 \cite{henderson2014second} and MultiWOZ2.1 \cite{eric2019multiwoz}. The statistics for all datasets are shown in Table \ref{datasets}. 

Sim-M and Sim-R are multi-turn dialogue datasets in the $movie$ and $restaurant$ domains, respectively, which are specially designed to evaluate the scalability of dialogue state tracking model. A large number of unknown slot values are included in their test set, so the generalization ability of the model can be reflected more accurately.

WOZ2.0 and DSTC2 datasets are both collected in the $restaurant$ domain and have the same three slots $food$, $area$, and $price\_range$. These two datasets provide automatic speech recognition (ASR) hypotheses of user utterances and can therefore be used to verify the robustness of the model against ASR errors. As in previous works, we use manuscript user utterance for training and top ASR hypothesis for testing.

MultiWOZ2.1 is the corrected version of the MultiWOZ \cite{budzianowski2018multiwoz}. Compared to the four datasets above, MultiWOZ2.1 is a more challenging and currently widely used benchmark for multi-turn multi-domain dialogue state tracking, consisting of 7 domains, over 30 slots, and over 4500 possible slot values. Following previous works \cite{wu-etal-2019-transferable,kim-etal-2020-efficient,heck2020trippy,zhu2020efficient}, we only use 5 domains ($restaurant$, $train$, $hotel$, $taxi$, $attraction$) that contain a total of 30 slots.

\subsection{Baselines}
We use 5 different types of baselines whose characteristics are shown in Table \ref{baselines}.

\paragraph{SpanPtr:}This is the first model to extract slot values directly from dialogue context without an ontology, it encodes the whole dialogue history with a bidirectional RNN and extracts slot value for each slot by generating the start and end positions in dialogue history \cite{xu2018end}.
\paragraph{TRADE:}This model is the first to consider knowledge transfer between domains in the multi-domain dialogue state tracking task. It represents a slot as a concatenation of domain name and slot name, encodes all dialogue history using bidirectional RNN, and finally decodes each slot value using a pointer-generator network \cite{wu-etal-2019-transferable}.
\paragraph{BERTDST:}This model decodes only the slot values of the slots mentioned in the current turn of dialogue, and then uses a rule-based update mechanism to update from the previous state to the current turn state. It uses BERT to encode the current turn of dialogue and extracts slot values from the dialogue as spans \cite{chao2019bert}.
\paragraph{SOMDST:}This model takes the dialogue state as an explicit memory that can be selectively overwritten, and inputs it into BERT together with the current turn dialogue. It then decomposes the prediction for each slot value into operation prediction and slot generation \cite{kim-etal-2020-efficient}.
\paragraph{SUMBT:}This model uses an ontology and is trained and evaluated on the dialogue session level instead of the dialogue turn level. BERT is used in the model to encode turn level dialogues, and an unidirectional RNN is used to capture session-level representation \cite{lee2019sumbt}.

\subsection{Configurations and Metrics}
Our deployments are based on the official implementation source code of SOMDST\footnote{\url{https://github.com/clovaai/som-dst}} and SUMBT\footnote{\url{https://github.com/SKTBrain/SUMBT}}, in which SpanPtr, TRADE and BERTDST are reproduced in this paper. BERT in all models uses pre-trained BERT \cite{vaswani2017attention} (BERT-Base, Uncased) which has 12 hidden layers of 768 units and 12 self-attention heads, while RNN uses GRU \cite{cho2014learning}. We use adam \cite{kingma2014adam} as the optimizer and use greedy decoding. We customize the training epochs for all models, and the training stopped early when the model's performance on development set failed to improve for 15 consecutive epochs, and all the results were averaged over the three runs with different random seeds. The detailed setting of the hyperparameters is given in Appendix A.

Since the length of the dialogue history is related to the granularity, the input length of the model needs to adapt to the granularity. Especially for the model with BERT as the encoder, in order to prevent the input from being truncated, we set the max sequence length to exceed almost all the inputs under different granularity. See Appendix A for details on the max sequence length settings. 

Following previous works \cite{xu2018end,wu-etal-2019-transferable,kim-etal-2020-efficient,heck2020trippy}, the joint accuracy (Joint acc) and slot accuracy (Slot acc) are used for evaluation. The joint accuracy is the accuracy that checks whether all the predicted slot values in each turn are exactly the same as the ground truth slot values. The slot accuracy is the average accuracy of slot value prediction in all turns.

\begin{table*}[!t]
	\centering
	\begin{tabular}{ccccccccc}
		\toprule
		\multirow{2}{*}{Models} & \multirow{2}{*}{TG} & \multirow{2}{*}{IG}  & \multicolumn{2}{c}{WOZ2.0} & \multicolumn{2}{c}{DSTC2} & \multicolumn{2}{c}{MultiWOZ2.1}  \\
		& & & Joint acc & Slot acc & Joint acc & Slot acc & Joint acc & Slot acc \\
		\toprule
		\multirow{4}{*}{SpanPtr} & 0* & 0* & 0.4455 & 0.7475 & \textbf{0.6234} & \textbf{0.8461} & \textbf{0.4415} & \textbf{0.9570} \\
		& -1 & -1 & 0.5012 & 0.7786 & 0.5829 & 0.8251 & 0.3868 & 0.9495 \\
		& -2 & -2 & 0.5881 & 0.8121 & 0.4825 & 0.7728 & 0.3726 & 0.9499 \\
		& -3 & -3 & \textbf{0.6330} & \textbf{0.8350} & 0.4737 & 0.7628 & 0.3745 & 0.9507 \\
		\hline
		\multirow{4}{*}{TRADE} & 0* & 0* & \textbf{0.5808} & \textbf{0.8186} & \textbf{0.6493} & \textbf{0.8590} & \textbf{0.4420} & \textbf{0.9655} \\
		& -1 & -1 & 0.5194 & 0.7833 & 0.5013 & 0.7834 & 0.3963 & 0.9613 \\
		& -2 & -2 & 0.5680 & 0.8107 & 0.4185 & 0.7488 & 0.3528 & 0.9569 \\
		& -3 & -3 & 0.5292 & 0.7886 & 0.5171 & 0.7963 & 0.3564 & 0.9552 \\
		\hline
		\multirow{4}{*}{BERTDST} & 1*& 1*& 0.8194 & 0.9307 & \textbf{0.6395} & \textbf{0.8537} & 0.4140 & 0.9584 \\
		& 2 & 2 & 0.8220 & 0.9318 & 0.5830 & 0.8271 & 0.4586 & 0.9636 \\
		& 3 & 3 & 0.8190 & 0.9318 & 0.5614 & 0.8103 & 0.4772 & 0.9646 \\
		& 4 & 4 & \textbf{0.8256} & \textbf{0.9344} & 0.5666 & 0.8152 & \textbf{0.4917} & \textbf{0.9659} \\
		\hline
		\multirow{4}{*}{SOMDST} & 1*& 1*& 0.8540 & 0.9471 & 0.6975 & 0.8828 & 0.5029 & 0.9715 \\
		& 2 & 2 & 0.8274 & 0.9341 & 0.7022 & 0.8808 & \textbf{0.5179} & \textbf{0.9730} \\
		& 3 & 3 & 0.8280 & 0.9356 & 0.7121 & 0.8851 & 0.5128 & 0.9720 \\
		& 4 & 4 & \textbf{0.8620} & \textbf{0.9491} & \textbf{0.7176} & \textbf{0.8882} & 0.5085 & 0.9718 \\
		\bottomrule
	\end{tabular}
	\caption{\label{table_leap_tracking} Joint accuracy and slot accuracy on WOZ2.0, DSTC2 and MultiWOZ2.1 when the same granularities are used in the training and inference phases. TG and IG are the training granularity and inference granularity, respectively. * refers to the granularity originally used in the baseline. }
\end{table*}

\section{Experimental Analysis}
This section presents our detailed investigation of how the context information of different granularity affects dialogue state tracking, focusing on the impact of granularity on dialogue state tracking, the combination of multiple granularities, and the application of context granularity in few-shot learning scenario. For simplicity, in all experimental results, the maximum granularity is expressed as 0, the maximum granularity minus 1 is expressed as -1, and so on.

\subsection{How greatly different granularities affect dialogue state tracking?}

The first part of our investigation look at the validity of the context granularity used by the current various dialogue state tracking models and try to figure out how different granularities affect dialogue state tracking. The experimental results are shown in Table \ref{table_leap_tracking}.

It can be found that some dialogue state tracking models do not take the appropriate granularity, and their performance is greatly improved when they are trained with the the context of appropriate granularity. For example, the joint accuracy of SpanPtr with granularity -3 on WOZ2.0 improved by 42\%, while the joint accuracy of BERTDST with granularity 4 on MultiWOZ2.1 improved by 19\%. These results suggest that there are significant differences in dialogue state tracking at different granularities, therefore, we should be careful to determine the granularity to be used according to the characteristics of the model and dataset.

By observing the experimental comparison results on different models and datasets in Table \ref{table_leap_tracking}, it can be found that:

\begin{figure}[!t]
	\centering
	\includegraphics[width=\linewidth]{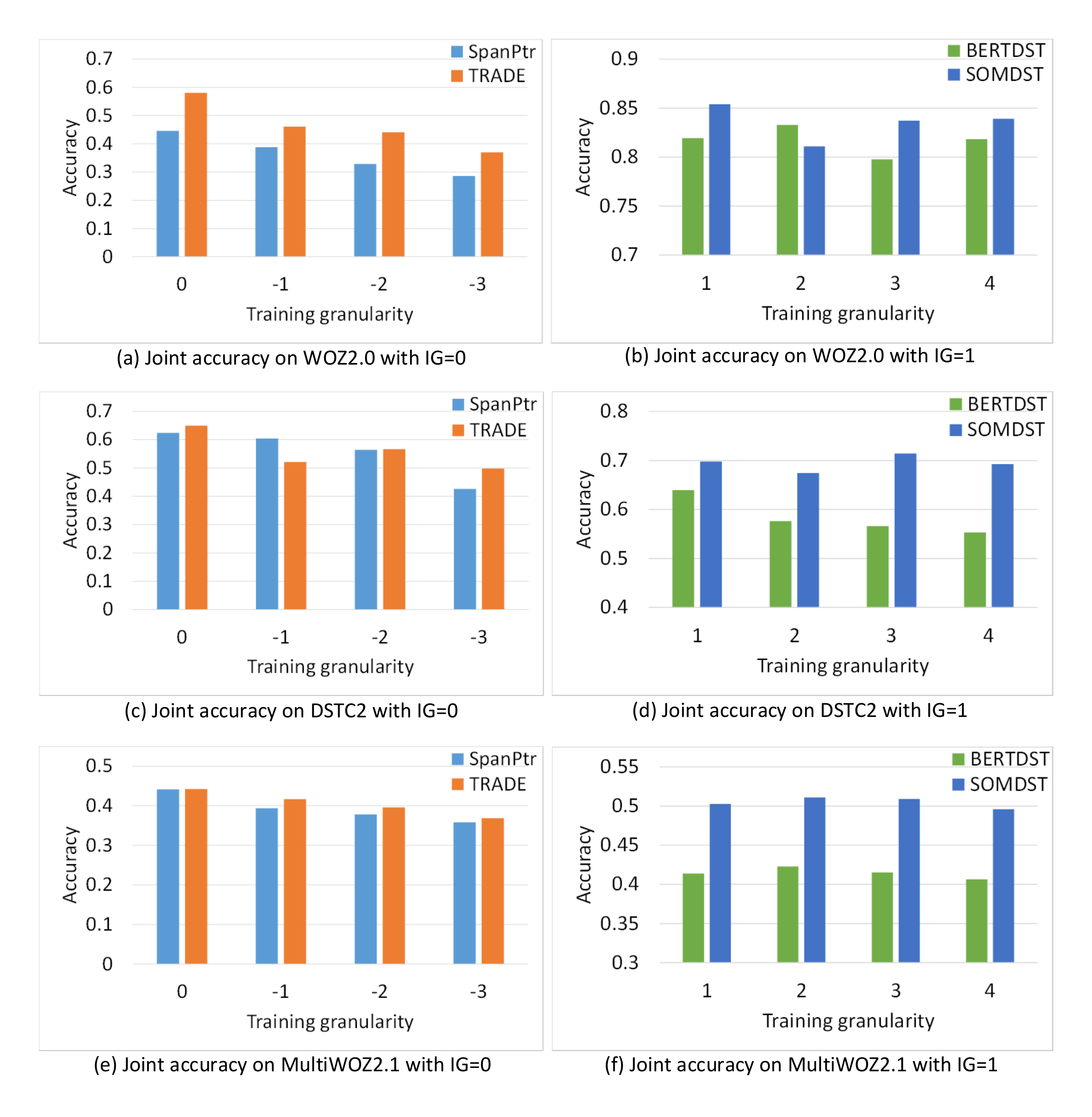}
	\caption{Joint accuracy of baseline model when context information with different granularity is used in training and inference phases. IG is the inference granularity.}
	\label{fig_leap_tracking}
\end{figure}

\begin{table*}[!t]
	\centering
	\begin{tabular}{ccccccccc}
		\toprule
		\multirow{2}{*}{Models} & \multirow{2}{*}{TG} & \multirow{2}{*}{IG} & \multicolumn{2}{c}{WOZ2.0} & \multicolumn{2}{c}{DSTC2} & \multicolumn{2}{c}{MultiWOZ2.1}  \\
		& & & Joint acc & Slot acc & Joint acc & Slot acc & Joint acc & Slot acc \\
		\toprule
		\multirow{2}{*}{SpanPtr}           & 0* & 0* & 0.4455 & \textbf{0.7475} & \textbf{0.6234} & \textbf{0.8461} & 0.4415 & \textbf{0.9570}  \\
		& {0, -1} & 0 & \textbf{0.4804} & 0.7428 & 0.6078 & 0.8371 & \textbf{0.4430} & 0.9565  \\
		\hline
		\multirow{2}{*}{TRADE}           & 0* & 0* & 0.5808 & 0.8186 & \textbf{0.6493} & \textbf{0.8590} & \textbf{0.4420} & \textbf{0.9655}  \\
		& {0, -1} & 0 & \textbf{0.6102} & \textbf{0.8357} & 0.6030 & 0.8413 & 0.4410 & \textbf{0.9655}  \\
		\hline
		\multirow{2}{*}{BERTDST}           & 1* & 1* & 0.8194 & 0.9307 & \textbf{0.6395} & \textbf{0.8537} & 0.4140 & 0.9584  \\
		& {1, 2} & 1 & \textbf{0.8331} & \textbf{0.9368} & 0.5824 & 0.8290 & \textbf{0.4229} & \textbf{0.9602}  \\
		\hline
		\multirow{2}{*}{SOMDST}           & 1* & 1* & 0.8540 & 0.9471 & 0.6975 & 0.8828 & 0.5029 & 0.9715  \\
		& {1, 2} & 1 & \textbf{0.8572} & \textbf{0.9479} & \textbf{0.7077} & \textbf{0.8866} & \textbf{0.5126} & \textbf{0.9723}  \\
		\hline
		\multirow{2}{*}{SUMBT}           & 1* & 1* & 0.9052 & 0.9665 & 0.6571 & 0.8664 & 0.4632 & 0.9655  \\
		& {1, 2} & 1 & \textbf{0.9089} & \textbf{0.9677} & \textbf{0.6739} & \textbf{0.8716} & \textbf{0.4725} & \textbf{0.9663}  \\
		\bottomrule
	\end{tabular}
	\caption{\label{table_MGL} Comparison of different baseline models on WOZ2.0, DSTC2 and MultiWOZ2.1 before and after applying multi-granularity combination. TG and IG are the training granularity and inference granularity, respectively. * refers to the granularity originally used in the baseline.}
\end{table*}

\begin{figure*}[!t]
	\centering
	\includegraphics[width=\linewidth]{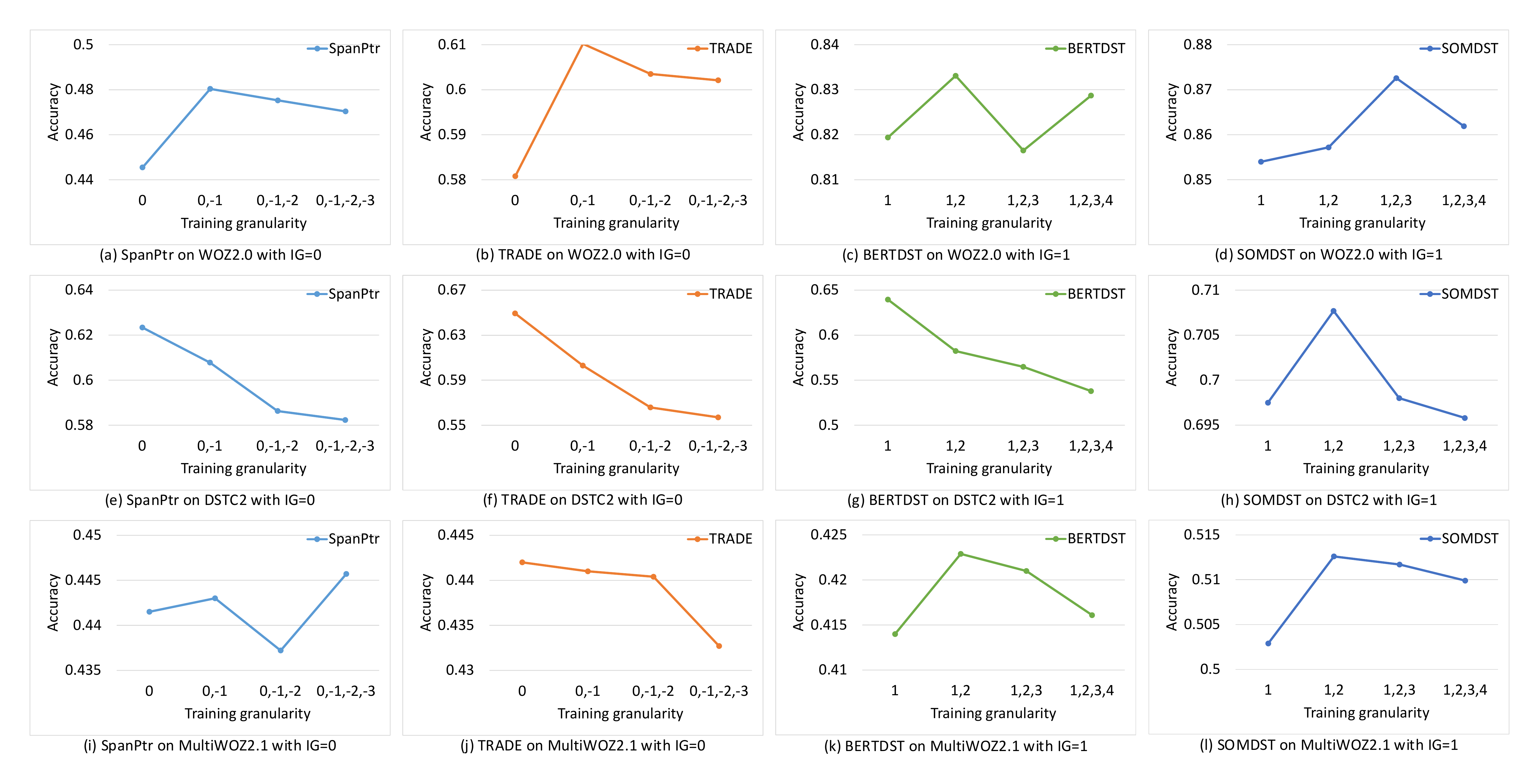}
	\caption{Joint accuracy of baseline model with different multi-granularity combinations is adopted in training phase. IG is the inference granularity.}
	\label{fig_MGL_result}
\end{figure*}

\begin{itemize}
	\item For different models, the model with generative decoding prefer larger granularity, because it requires more context information to effectively learn vocabulary-based distribution. For example, TRADE and SOMDST both perform better in larger granularity. Meanwhile, the model with extractive decoding is more dependent on the characteristics of the dataset. Besides, in general, the model with generative decoding has obvious advantages over the model with extractive decoding.
	\item For different datasets, when the dataset involves multiple domains and there are a large number of long-dependency dialogue states, context information of larger granularity can be used to more effectively capture the long-dependency relationship in the data for dialogue state tracking, such as MultiWOZ2.1 dataset. For simpler single-domain datasets, where a large number of short dependencies determine the effectiveness of small granularity in dialogue state tracking. However, when there are more turns of dialogue resulting in less information in each turn, a larger granularity may be required to provide enough information, for example, SpanPtr performs best on the DSTC2 dataset at maximum granularity.
\end{itemize}

As can be seen from the above analysis, different granularities have their own advantages in different situations of dialogue, so it is natural to wonder whether multiple granularities can be combined to achieve better dialogue state tracking. Next, let's discuss the issue of multi-granularity combination.

\subsection{How to combine multiple granularities for dialogue state tracking?}

Following the above analysis, here we mainly discuss how to combine multiple granularities in dialogue state tracking, mainly focusing on three aspects: (1) The relationship between granularities, (2) Performance of multi-granularity combination and (3) Limitations of multi-granularity combination.

\paragraph{The relationship between granularities:}
First, we use different granularities in the training and inference phases of dialogue state tracking to figure out the relationship between different granularities, as shown in Figure \ref{fig_leap_tracking}. It can be seen that when we fix the granularity of context information in the inference phase, the dialogue state tracking model trained with other granularity still obtains the generalization under this inference granularity. And even some models learned at other granularity, such as the BERTDST in Figure \ref{fig_leap_tracking} (b) and (f), can perform better. Meanwhile, it can also be found that as the granularity gap increases, the context information becomes more and more inconsistent, and eventually the ability of the model to generalize across granularity is gradually reduced. Through these phenomena, we can summarize as follows: The knowledge learned by the dialogue state tracking model in context information of different granularity is transferable and the smaller the gap between granularity can bring more knowledge transfer effect.

\paragraph{Performance of multi-granularity combination:}
Then, we use the knowledge transfer between context information of different granularity to improve the baseline. In the specific experiment, we add the most adjacent granularity to the training phase of the model, that is, the context under two granularities is used for training, while the inference phase remains unchanged, as shown in Table \ref{table_MGL}. It can be observed that in most cases, the performance of the baseline models is significantly enhanced, suggesting that adding more granularity context information to the training phase of the model can indeed  improve the generalization of the dialogue state tracking model. Of course, in some cases, multi-granularity combination results in a reduction in performance, such as SpanPtr, TRADE, and BERTDST on DSTC2 dataset. The main reason for this phenomenon should be the large deviation between the context information of different granularity in the multi-granularity combination, as can be seen from the large reduction of SpanPtr, TRADE, and BERTDST on the DSTC2 dataset with other granularity in Table \ref{table_leap_tracking}.

\paragraph{Limitations of multi-granularity combination:}
Given that multi-granularity combination can lead to improved generalization performance, is it better to have more context information of different granularity in training phase? To answer this question, we gradually add more granularities to the training phase while keeping the inference granularity unchanged, the experimental results are shown in Figure \ref{fig_MGL_result}. It can be found that there is an upper limit to the use of multi-granularity combination in the training phase. Generally, adding the granularity with the smallest gap can bring the best effect, after that, with the increase of granularity number, the performance will decline.

\subsection{Application of context information granularity in few-shot learning scenario}

\begin{figure}[!t]
	\centering
	\includegraphics[width=\linewidth]{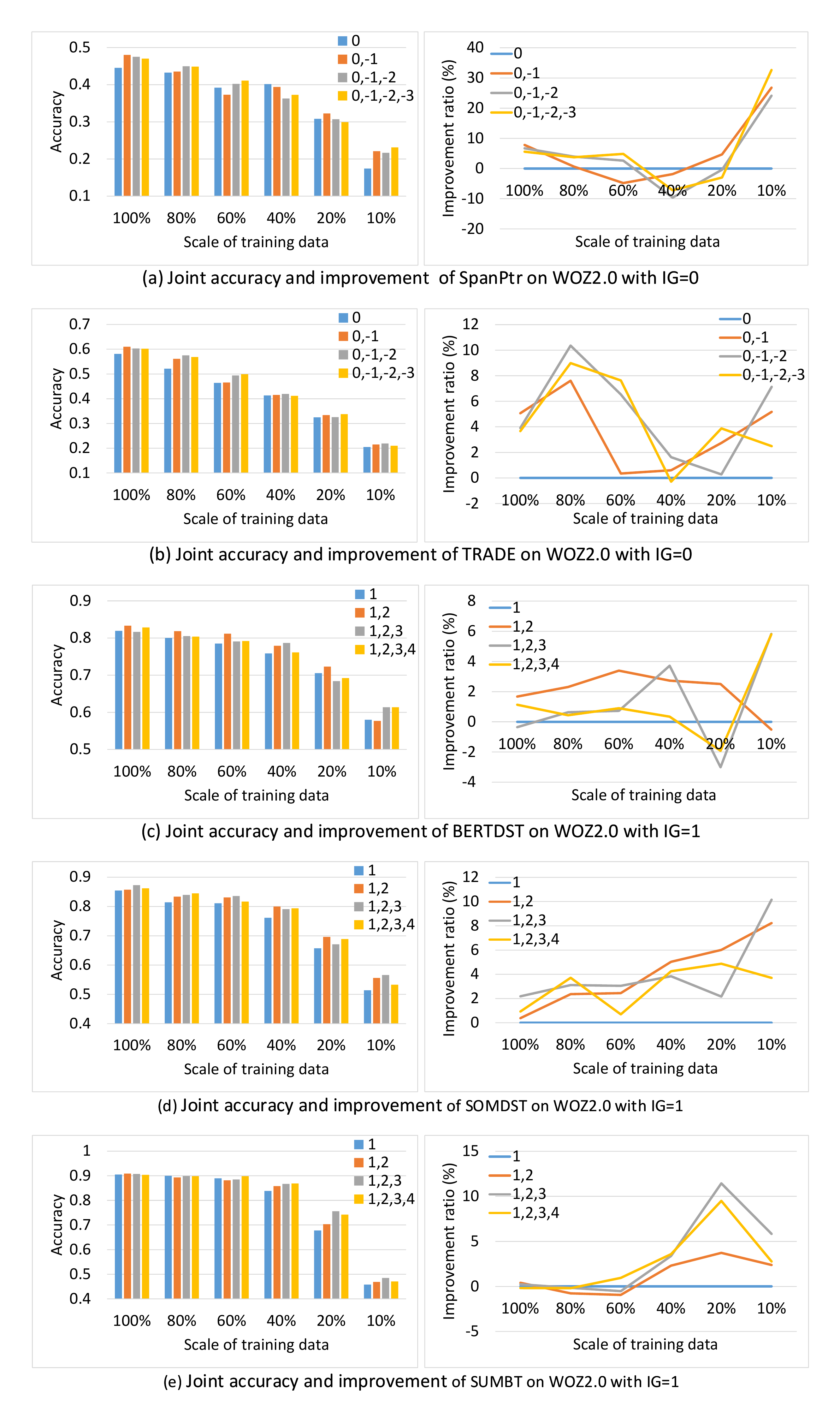}
	\caption{Joint accuracy of baseline model and improvement ratio of multi-granularity combination under different scales of training data. Items in different colors represent different granularity combinations in training phase, and IG is the inference granularity.}
	\label{data_scale}
\end{figure}

\begin{table*}[!t]
	\centering
	\begin{tabular}{cccccccc}
		\toprule
		\multirow{2}{*}{Models} & \multirow{2}{*}{TG} & \multirow{2}{*}{IG} & Sim-M & Sim-R & WOZ2.0 & DSTC2 & MultiWOZ2.1 \\
		& & & 10\% & 10\% & 10\% & 5\% & 5\% \\
		\toprule
		\multirow{4}{*}{SpanPtr} & 0* & 0* & \textbf{0.1466} & 0.5147 & 0.1744 & 0.4523 & 0.2700  \\
		& {0, -1} & 0         & 0.1188 & 0.5631 & 0.2211 & 0.4640 & 0.2703 \\
		& {0, -1, -2} & 0     & 0.0985 & 0.5752 & 0.2165 & 0.4839 & \textbf{0.2765}  \\
		& {0, -1, -2, -3} & 0 & 0.0872 & \textbf{0.5805} & \textbf{0.2313} & \textbf{0.4873} & 0.2762  \\
		\hline
		\multirow{4}{*}{TRADE} & 0* & 0* & 0.0780 & 0.6531 & 0.2047 & \textbf{0.5290} & \textbf{0.2531}  \\
		& {0, -1} & 0         & 0.0880 & 0.6512 & 0.2153 & 0.5108 & 0.2400  \\
		& {0, -1, -2} & 0     & 0.0892 & \textbf{0.6612} & \textbf{0.2193} & 0.5173 & 0.2470    \\
		& {0, -1, -2, -3} & 0 & \textbf{0.0921} & 0.6569 & 0.2098 & 0.5101 & 0.2461    \\
		\hline
		\multirow{4}{*}{BERTDST} & 1* & 1* & 0.4814 & 0.7066 & 0.5800 & 0.4697 & 0.3414  \\
		& {1, 2} & 1          & \textbf{0.6219} & 0.7295 & 0.5770 & \textbf{0.5137} & \textbf{0.3491}  \\
		& {1, 2, 3} & 1       & 0.5926 & \textbf{0.7376} & \textbf{0.6138} & 0.4712 & 0.3450   \\
		& {1, 2, 3, 4} & 1    & 0.6075 & 0.7241 & 0.6136 & 0.4929 & 0.3377   \\
		\hline
		\multirow{4}{*}{SOMDST} & 1* & 1* & 0.2708 & 0.4700 & 0.5140 & 0.3967 & 0.3596  \\
		& {1, 2} & 1          & \textbf{0.2754} & 0.5101 & 0.5563 & 0.5151 & \textbf{0.3706}  \\
		& {1, 2, 3} & 1       & 0.2549 & \textbf{0.5166} & \textbf{0.5662} & \textbf{0.5307} & 0.3613  \\
		& {1, 2, 3, 4} & 1    & 0.2104 & 0.5142 & 0.5330 & 0.5238 & 0.3572   \\
		\hline
		\multirow{4}{*}{SUMBT} & 1* & 1* & \textbf{0.0982} & 0.6526 & 0.4581 & 0.4689 & 0.2964  \\
		& {1, 2} & 1          & 0.0980 & \textbf{0.6546} & 0.4690 & 0.5493 & 0.3535  \\
		& {1, 2, 3} & 1       & 0.0980 & 0.6390 & \textbf{0.4848} & 0.5265 & \textbf{0.3696}  \\
		& {1, 2, 3, 4} & 1    & 0.0968 & 0.6464 & 0.4708 & \textbf{0.5611} & 0.3637  \\
		\bottomrule
	\end{tabular}
	\caption{\label{table_few_shot} Joint accuracy of baseline models in few-shot learning before and after applying multi-granularity combination in training phase. TG and IG are the training granularity and inference granularity, respectively. * refers to the granularity originally used in the baseline. 10\% and 5\% refer to the scale of the training data.}
\end{table*}

Considering the knowledge transfer between granularity in multi-granularity combination, we explore the application of multi-granularity combination in few-shot learning scenario.

Figure \ref{data_scale} shows the joint accuracy of the model with different multi-granularity combinations and the percentage improvement relative to the baseline model on the WOZ2.0 dataset with different training data scales. It can be found that under different scales of training data, multi-granularity combination can achieve better performance compared with single-granularity in most cases. Moreover, it can be seen from (a), (d) and (e) that the advantages of multi-granularity combination are gradually expanding with the decrease of the scale of training dataset. Therefore, the performance of multi-granularity combination in few-shot learning is worth exploring.

We conduct detailed experiments on all the 5 datasets in the paper to fully explore the potential of multi-granularity combination in few-shot learning, as shown in Table \ref{table_few_shot}. It can be found that multi-granularity combination has a very significant effect in few-shot learning, and in some cases can even achieve a relative improvement of more than 10\%, such as SpanPtr on Sim-R and WOZ2.0, BERTDST on Sim-M, SOMDST on WOZ2.0 and DSTC2. Meanwhile, in few-shot learning, the upper limit of multi-granularity combination can be higher, and better performance can be achieved when more granularities are added in the training phase.

The above experimental results of multi-granularity combination in few-shot learning show that, there is indeed knowledge transfer between different granularity contexts, and the model can obtain more adequate modeling of dialogue by learning context dialogues of different granularity.

\section{Conclusion}
In the paper, we analyze the defects of two existing traditional dialogue state tracking strategies when dealing with context of different granularity and make a comprehensive study on how the context information of different granularity affects dialogue state tracking. Extensive experimental results and analysis show that: (1) Different granularities have their own advantages in different situations of dialogue state tracking; (2) The multi-granularity combination can effectively improve the dialogue state tracking; (3) The application of multi-granularity combination in few-shot learning can bring significant effects. In future work, dynamic context granularity can be used in training and inference to further improve dialogue state tracking.

\section{Ethical Consideration}
This work may contribute to the development of conversational systems. In the narrow sense, this work focuses on dialogue state tracking in task-oriented dialogue system, hoping to improve the ability of conversational AI to understand human natural language. If so, these improvements could have a positive impact on the research and application of conversational AI, which could help humans to complete goals more effectively in a more intelligent way of communication. However, we never forget the other side of the coin. The agent substitution of conversational AI may affect the humanized communication and may lead to human-machine conflict problems, which need to be considered more broadly in the field of conversational AI.

\section*{Acknowledgments}
We thank the anonymous reviewers for their insightful comments. This work was supported by National Key R\&D Plan (No. 2020AAA0106600) and National Natural Science Foundation of China (Grant No. U19B2020 and No. 61772076).

\bibliographystyle{acl_natbib}
\bibliography{references}

\begin{thebibliography}{25}
\expandafter\ifx\csname natexlab\endcsname\relax\def\natexlab#1{#1}\fi

\bibitem[{Budzianowski et~al.(2018)Budzianowski, Wen, Tseng, Casanueva, Ultes,
  Ramadan, and Gasic}]{budzianowski2018multiwoz}
Pawe{\l} Budzianowski, Tsung-Hsien Wen, Bo-Hsiang Tseng, I{\~n}igo Casanueva,
  Stefan Ultes, Osman Ramadan, and Milica Gasic. 2018.
\newblock Multiwoz-a large-scale multi-domain wizard-of-oz dataset for
  task-oriented dialogue modelling.
\newblock In \emph{Proceedings of the 2018 Conference on Empirical Methods in
  Natural Language Processing}, pages 5016--5026.

\bibitem[{Chao and Lane(2019)}]{chao2019bert}
Guan-Lin Chao and Ian Lane. 2019.
\newblock Bert-dst: Scalable end-to-end dialogue state tracking with
  bidirectional encoder representations from transformer.
\newblock \emph{Proc. Interspeech 2019}, pages 1468--1472.

\bibitem[{Chen et~al.(2017)Chen, Liu, Yin, and Tang}]{chen2017survey}
Hongshen Chen, Xiaorui Liu, Dawei Yin, and Jiliang Tang. 2017.
\newblock A survey on dialogue systems: Recent advances and new frontiers.
\newblock \emph{Acm Sigkdd Explorations Newsletter}, 19(2):25--35.

\bibitem[{Cho et~al.(2014)Cho, van Merrienboer, Gulcehre, Bougares, Schwenk,
  and Bengio}]{cho2014learning}
Kyunghyun Cho, B~van Merrienboer, Caglar Gulcehre, F~Bougares, H~Schwenk, and
  Yoshua Bengio. 2014.
\newblock Learning phrase representations using rnn encoder-decoder for
  statistical machine translation.
\newblock In \emph{Conference on Empirical Methods in Natural Language
  Processing (EMNLP 2014)}.

\bibitem[{Eric et~al.(2019)Eric, Goel, Paul, Sethi, Agarwal, Gao, and
  Hakkani-Tur}]{eric2019multiwoz}
Mihail Eric, Rahul Goel, Shachi Paul, Abhishek Sethi, Sanchit Agarwal, Shuyag
  Gao, and Dilek Hakkani-Tur. 2019.
\newblock Multiwoz 2.1: Multi-domain dialogue state corrections and state
  tracking baselines.
\newblock \emph{arXiv preprint arXiv:1907.01669}.

\bibitem[{Goel et~al.(2018)Goel, Paul, Chung, Lecomte, Mandal, and
  Hakkani-Tur}]{goel2018flexible}
Rahul Goel, Shachi Paul, Tagyoung Chung, Jeremie Lecomte, Arindam Mandal, and
  Dilek Hakkani-Tur. 2018.
\newblock Flexible and scalable state tracking framework for goal-oriented
  dialogue systems.
\newblock \emph{arXiv preprint arXiv:1811.12891}.

\bibitem[{Goel et~al.(2019)Goel, Paul, and Hakkani-T{\'u}r}]{goel2019hyst}
Rahul Goel, Shachi Paul, and Dilek Hakkani-T{\'u}r. 2019.
\newblock Hyst: A hybrid approach for flexible and accurate dialogue state
  tracking.
\newblock \emph{Proc. Interspeech 2019}, pages 1458--1462.

\bibitem[{Heck et~al.(2020)Heck, van Niekerk, Lubis, Geishauser, Lin, Moresi,
  and Gasic}]{heck2020trippy}
Michael Heck, Carel van Niekerk, Nurul Lubis, Christian Geishauser, Hsien-Chin
  Lin, Marco Moresi, and Milica Gasic. 2020.
\newblock Trippy: A triple copy strategy for value independent neural dialog
  state tracking.
\newblock In \emph{Proceedings of the 21th Annual Meeting of the Special
  Interest Group on Discourse and Dialogue}, pages 35--44.

\bibitem[{Henderson et~al.(2014)Henderson, Thomson, and
  Williams}]{henderson2014second}
Matthew Henderson, Blaise Thomson, and Jason~D Williams. 2014.
\newblock The second dialog state tracking challenge.
\newblock In \emph{Proceedings of the 15th annual meeting of the special
  interest group on discourse and dialogue (SIGDIAL)}, pages 263--272.

\bibitem[{Kim et~al.(2020)Kim, Yang, Kim, and Lee}]{kim-etal-2020-efficient}
Sungdong Kim, Sohee Yang, Gyuwan Kim, and Sang-Woo Lee. 2020.
\newblock Efficient dialogue state tracking by selectively overwriting memory.
\newblock In \emph{Proceedings of the 58th Annual Meeting of the Association
  for Computational Linguistics}, pages 567--582.

\bibitem[{Kingma and Ba(2014)}]{kingma2014adam}
Diederik~P Kingma and Jimmy Ba. 2014.
\newblock Adam: A method for stochastic optimization.
\newblock \emph{arXiv preprint arXiv:1412.6980}.

\bibitem[{Lee et~al.(2019)Lee, Lee, and Kim}]{lee2019sumbt}
Hwaran Lee, Jinsik Lee, and Tae-Yoon Kim. 2019.
\newblock Sumbt: Slot-utterance matching for universal and scalable belief
  tracking.
\newblock In \emph{Proceedings of the 57th Annual Meeting of the Association
  for Computational Linguistics}, pages 5478--5483.

\bibitem[{Lei et~al.(2018)Lei, Jin, Kan, Ren, He, and Yin}]{lei2018sequicity}
Wenqiang Lei, Xisen Jin, Min-Yen Kan, Zhaochun Ren, Xiangnan He, and Dawei Yin.
  2018.
\newblock Sequicity: Simplifying task-oriented dialogue systems with single
  sequence-to-sequence architectures.
\newblock In \emph{Proceedings of the 56th Annual Meeting of the Association
  for Computational Linguistics (Volume 1: Long Papers)}, pages 1437--1447.

\bibitem[{Mrk{\v{s}}i{\'c} et~al.(2017)Mrk{\v{s}}i{\'c}, S{\'e}aghdha, Wen,
  Thomson, and Young}]{mrkvsic2017neural}
Nikola Mrk{\v{s}}i{\'c}, Diarmuid~{\'O} S{\'e}aghdha, Tsung-Hsien Wen, Blaise
  Thomson, and Steve Young. 2017.
\newblock Neural belief tracker: Data-driven dialogue state tracking.
\newblock In \emph{Proceedings of the 55th Annual Meeting of the Association
  for Computational Linguistics (Volume 1: Long Papers)}, pages 1777--1788.

\bibitem[{Rastogi et~al.(2018)Rastogi, Gupta, and
  Hakkani-Tur}]{rastogi2018multi}
Abhinav Rastogi, Raghav Gupta, and Dilek Hakkani-Tur. 2018.
\newblock Multi-task learning for joint language understanding and dialogue
  state tracking.
\newblock In \emph{Proceedings of the 19th Annual SIGdial Meeting on Discourse
  and Dialogue}, pages 376--384.

\bibitem[{Rastogi et~al.(2017)Rastogi, Hakkani-T{\"u}r, and
  Heck}]{rastogi2017scalable}
Abhinav Rastogi, Dilek Hakkani-T{\"u}r, and Larry Heck. 2017.
\newblock Scalable multi-domain dialogue state tracking.
\newblock In \emph{2017 IEEE Automatic Speech Recognition and Understanding
  Workshop (ASRU)}, pages 561--568. IEEE.

\bibitem[{Ren et~al.(2019)Ren, Ni, and McAuley}]{ren2019scalable}
Liliang Ren, Jianmo Ni, and Julian McAuley. 2019.
\newblock Scalable and accurate dialogue state tracking via hierarchical
  sequence generation.
\newblock In \emph{Proceedings of the 2019 Conference on Empirical Methods in
  Natural Language Processing and the 9th International Joint Conference on
  Natural Language Processing (EMNLP-IJCNLP)}, pages 1876--1885.

\bibitem[{Shah et~al.(2018)Shah, Hakkani-T{\"u}r, T{\"u}r, Rastogi, Bapna,
  Nayak, and Heck}]{shah2018building}
Pararth Shah, Dilek Hakkani-T{\"u}r, Gokhan T{\"u}r, Abhinav Rastogi, Ankur
  Bapna, Neha Nayak, and Larry Heck. 2018.
\newblock Building a conversational agent overnight with dialogue self-play.
\newblock \emph{arXiv preprint arXiv:1801.04871}.

\bibitem[{Shan et~al.(2020)Shan, Li, Zhang, Meng, Feng, Niu, and
  Zhou}]{shan2020contextual}
Yong Shan, Zekang Li, Jinchao Zhang, Fandong Meng, Yang Feng, Cheng Niu, and
  Jie Zhou. 2020.
\newblock A contextual hierarchical attention network with adaptive objective
  for dialogue state tracking.
\newblock In \emph{Proceedings of the 58th Annual Meeting of the Association
  for Computational Linguistics}, pages 6322--6333.

\bibitem[{Vaswani et~al.(2017)Vaswani, Shazeer, Parmar, Uszkoreit, Jones,
  Gomez, Kaiser, and Polosukhin}]{vaswani2017attention}
Ashish Vaswani, Noam Shazeer, Niki Parmar, Jakob Uszkoreit, Llion Jones,
  Aidan~N Gomez, {\L}ukasz Kaiser, and Illia Polosukhin. 2017.
\newblock Attention is all you need.
\newblock In \emph{Proceedings of the 31st International Conference on Neural
  Information Processing Systems}, pages 6000--6010.

\bibitem[{Wen et~al.(2016)Wen, Vandyke, Mrksic, Gasic, Rojas-Barahona, Su,
  Ultes, and Young}]{wen2016network}
Tsung-Hsien Wen, David Vandyke, Nikola Mrksic, Milica Gasic, Lina~M
  Rojas-Barahona, Pei-Hao Su, Stefan Ultes, and Steve Young. 2016.
\newblock A network-based end-to-end trainable task-oriented dialogue system.
\newblock \emph{arXiv preprint arXiv:1604.04562}.

\bibitem[{Wu et~al.(2019)Wu, Madotto, Hosseini-Asl, Xiong, Socher, and
  Fung}]{wu-etal-2019-transferable}
Chien-Sheng Wu, Andrea Madotto, Ehsan Hosseini-Asl, Caiming Xiong, Richard
  Socher, and Pascale Fung. 2019.
\newblock Transferable multi-domain state generator for task-oriented dialogue
  systems.
\newblock In \emph{Proceedings of the 57th Annual Meeting of the Association
  for Computational Linguistics}, pages 808--819.

\bibitem[{Xu and Hu(2018)}]{xu2018end}
Puyang Xu and Qi~Hu. 2018.
\newblock An end-to-end approach for handling unknown slot values in dialogue
  state tracking.
\newblock In \emph{Proceedings of the 56th Annual Meeting of the Association
  for Computational Linguistics (Volume 1: Long Papers)}, pages 1448--1457.

\bibitem[{Zhang et~al.(2020)Zhang, Hashimoto, Wu, Wang, Philip, Socher, and
  Xiong}]{zhang2020find}
Jianguo Zhang, Kazuma Hashimoto, Chien-Sheng Wu, Yao Wang, S~Yu Philip, Richard
  Socher, and Caiming Xiong. 2020.
\newblock Find or classify? dual strategy for slot-value predictions on
  multi-domain dialog state tracking.
\newblock In \emph{Proceedings of the Ninth Joint Conference on Lexical and
  Computational Semantics}, pages 154--167.

\bibitem[{Zhu et~al.(2020)Zhu, Li, Chen, and Yu}]{zhu2020efficient}
Su~Zhu, Jieyu Li, Lu~Chen, and Kai Yu. 2020.
\newblock Efficient context and schema fusion networks for multi-domain
  dialogue state tracking.
\newblock In \emph{Proceedings of the 2020 Conference on Empirical Methods in
  Natural Language Processing: Findings}, pages 766--781.

\end{thebibliography}

\appendix
\onecolumn
\begin{appendices}
\section{Settings}

\begin{table}[h]
	\centering
	\begin{tabular}{cccccc}
		\toprule
		Hyperparameters & SpanPtr & TRADE & BERTDST & SOMDST & SUMBT \\
		\toprule
		Batch size & 32 & 32 & 16 & 16 & 4 \\
		Training epochs & 100 & 100 & 200 & 200 & 300 \\
		Early stop evaluation & Joint acc & Joint acc & Joint acc & Joint acc & Loss \\
		Decoder teacher forcing & - & 0.5 & - & 0.5 & - \\
		Dropout & 0.1 & 0.1 & 0.1 & 0.1 & - \\
		Word dropout & 0.1 & 0.1 & 0.1 & 0.1 & - \\
		RNN hidden size & 400 & 400 & 768 & 768 & 300 \\
		\multirow{2}{*}{Learning rate} & \multirow{2}{*}{1e-4} & \multirow{2}{*}{1e-3} & Enc: 4e-5 & Enc: 4e-5 & \multirow{2}{*}{5e-5}\\
		& & & Dec: 1e-4 & Dec: 1e-4 & \\
		Warmup proportion & - & - & 0.1 & 0.1 & 0.1 \\

		\bottomrule
	\end{tabular}
	\caption{\label{hyperparameters} The detailed setting of hyperparameters. $word\ dropout$ means to randomly replace the input tokens with the special [UNK] with a certain probability.}
\end{table}

\begin{table}[h]
	\centering
	\begin{tabular}{ccccccc}
		\toprule
		Models & TG & Sim-M & Sim-R & WOZ2.0 & DSTC2 & MultiWOZ2.1  \\
		\toprule
		SpanPtr & - & - & - & - & - & - \\
		\hline
		TRADE   & - & - & - & - & - & - \\
		\hline
		\multirow{4}{*}{BERTDST} & 1 & 70  & 70  & 100 & 60  & 100 \\
								 & 2 & 90  & 120 & 150 & 80  & 150 \\
								 & 3 & 120 & 150 & 170 & 110 & 210 \\
								 & 4 & 150 & 180 & 200 & 140 & 260 \\
		\hline
		\multirow{4}{*}{SOMDST}  & 1 & 120 & 120 & 120 & 70  & 320 \\
								 & 2 & 150 & 150 & 150 & 100 & 360 \\
								 & 3 & 180 & 190 & 190 & 130 & 410 \\
								 & 4 & 200 & 220 & 220 & 160 & 460 \\
		\hline
		\multirow{4}{*}{SUMBT}  & 1 & 60  & 70  & 100 & 50  & 100 \\
								& 2 & 90  & 120 & 120 & 70  & 150 \\
								& 3 & 120 & 140 & 140 & 100 & 210 \\
								& 4 & 130 & 160 & 160 & 120 & 260 \\
		\bottomrule
	\end{tabular}
	\caption{\label{table_max_len} The setting of the max sequence length of BERT in encoders of different models. To minimize truncation of the input, the max sequence length exceeds the length of almost all input sequences in the dataset. SpanPtr and TRADE use GRU as encoders. TG is the training granularity.}
\end{table}
\end{appendices}
\end{document}